\documentclass[10pt,twocolumn,letterpaper]{article}

\usepackage{iccv}              

\newcommand{\layerbench}{\textbf{LayerBench}}
\usepackage{multirow}

\definecolor{iccvblue}{rgb}{0.21,0.49,0.74}
\usepackage[pagebackref,breaklinks,colorlinks,allcolors=iccvblue]{hyperref}

\def\paperID{2346} 
\def\confName{ICCV}
\def\confYear{2025}

\title{Trans-Adapter: A Plug-and-Play Framework for Transparent Image Inpainting}

\author{
Yuekun Dai$\,\,\,\,$ Haitian Li $\,\,\,\,$ Shangchen Zhou $\,\,\, $ Chen Change Loy \\
S-Lab, Nanyang Technological University \\
\texttt{\small \{ydai005, liha0032, s200094, ccloy\}@ntu.edu.sg}\\ \vspace{-6mm}
{\tt\small \url{https://ykdai.github.io/projects/trans-adapter}}
}

\begin{document}

\maketitle
\begin{abstract}

RGBA images, with the additional alpha channel, are crucial for any application that needs blending, masking, or transparency effects, making them more versatile than standard RGB images. Nevertheless, existing image inpainting methods are designed exclusively for RGB images. Conventional approaches to transparent image inpainting typically involve placing a background underneath RGBA images and employing a two-stage process: image inpainting followed by image matting. This pipeline, however, struggles to preserve transparency consistency in edited regions, and matting can introduce jagged edges along transparency boundaries. To address these challenges, we propose \textbf{Trans-Adapter}, a plug-and-play adapter that enables diffusion-based inpainting models to process transparent images directly. Trans-Adapter also supports controllable editing via ControlNet and can be seamlessly integrated into various community models. To evaluate our method, we introduce LayerBench, along with a novel non-reference alpha edge quality evaluation metric for assessing transparency edge quality. We conduct extensive experiments on LayerBench to demonstrate the effectiveness of our approach.

\end{abstract}    
\section{Introduction}
\label{sec:intro}
\begin{figure}[t]
  \centering
   \includegraphics[width=1.0\linewidth]{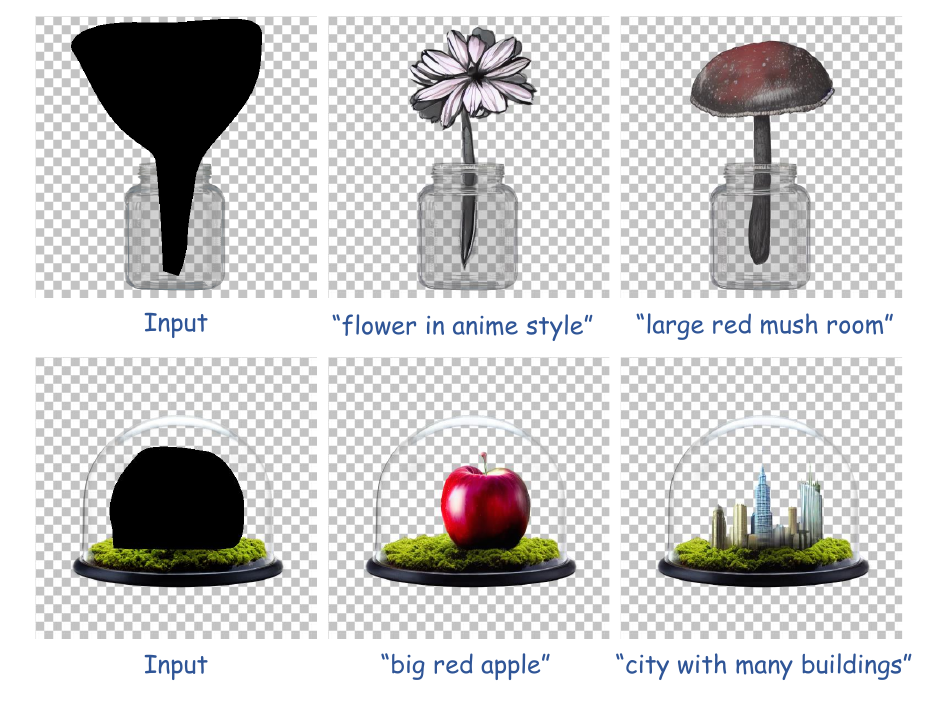}
   \vspace{-8mm}
   \caption{We introduce the first transparent image inpainting method. Given a transparent image, our approach can generate RGBA content within the masked region (shown in black) based on a provided text prompt.}
   \vspace{-3.5mm}
   \label{fig:trans_inpaint}
\end{figure}

Transparent images, typically represented as RGBA images, are widely used in film, animation, and game production.
These images are layered using various compositing techniques to produce the final scene.
To support this workflow, most commercial image and video editing software, such as Adobe Photoshop, Adobe After Effects, Live2D, and DragonBones, provide advanced layer management and image editing tools. These tools enable artists to manipulate individual layers with precision using brushes, pencils, and erasers, ensuring seamless integration of elements within a scene.
Despite the demand for editing transparent layers, AI-powered inpainting tools like generative fill remain limited to RGB images.
This limitation forces artists to manually refine or restore incomplete or unsatisfactory transparent images, often requiring tedious hand-painting to maintain visual coherence.

Image inpainting has seen remarkable improvements with the advent of text-to-image (T2I) diffusion models. These image inpainting models are typically fine-tuned from foundation generative models~\cite{sd15,sdxl,flux}. Many studies~\cite{ju2024brushnet,burgert2024magick,manukyan2023hdpaint,powerpaint,zhang2023controlnet} have further enhanced these models by introducing flexible user control through semantic and structural conditions. 
However, these methods are designed specifically for RGB images. While some models can inpaint the alpha and RGB channels of transparent images separately, inconsistencies between the generated content in these channels often lead to misalignment, resulting in unnatural inpainting artifacts.

In this study, we extend the functionality of inpainting to transparent images by proposing \textbf{Trans-Adapter}, a plug-and-play adapter that learns to inpaint aligned RGB and alpha content. 
Trans-Adapter enables transparent image editing without compromising inpainting performance, allowing seamless integration with existing diffusion-based inpainting models.
We decompose the RGBA image into separate RGB and alpha channels, treating them as a two-frame video. We then inflate a T2I model to process this ``video'' by incorporating a spatial alignment module and cross-domain self-attention, ensuring alignment and geometric consistency between the RGB and alpha channels. With this design, our method effectively preserves structural coherence across transparent regions, enabling seamless and high-quality inpainting results (Fig.~\ref{fig:trans_inpaint}). 
To train the adapter, we collect a new dataset of 35K high-quality transparent images from online PNG stock sources. We merge this dataset with a subset of MAGICK~\cite{burgert2024magick}, a generated dataset of objects with high-quality alpha mattes.

To evaluate the effectiveness of our approach comprehensively, we introduce \layerbench, a new benchmark consisting of 800 transparent images. This dataset includes 400 natural images sourced from online PNG stocks and test images of previous image matting datasets~\cite{xu2017deep}, while the remaining 400 generated images are carefully selected from LayerDiffusion~\cite{zhang2024transparent} and MAGICK~\cite{burgert2024magick} to ensure diversity.
LayerBench differs significantly from existing RGB image inpainting benchmarks, such as EditBench~\cite{wang2023imagen} and BrushBench~\cite{ju2024brushnet} that do not involve alpha channels.
LayerBench is specifically constructed to evaluate the quality of transparent image inpainting, with a focus on RGB-alpha alignment.
We ensure that a significant portion of the inpainting masks overlap with the boundaries of the alpha channel, where transparency transitions occur. This design allows for a more rigorous assessment of how well methods preserve edge consistency between RGB and alpha channels.

When the RGB channels of a transparent image are not perfectly aligned with its alpha map, jagged or rough edges may emerge, an artifact commonly seen in the outputs of image segmentation and matting. These imperfections can degrade visual quality, especially when blending the transparent image with other elements in a composition.
To assess the misalignment between the RGB and alpha channels, we propose a new non-reference metric for evaluating alpha edge quality. This metric can be computed with our specially trained \textbf{Alpha Edge Quality (AEQ)} classifier.

Our contributions are as follows:
1) We propose a plug-and-play framework for transparent image inpainting, marking the first dedicated approach for this task.
2) We present LayerBench, a carefully curated benchmark specifically designed to evaluate transparent image inpainting methods. 
3) To complement LayerBench, we propose a new non-reference metric that quantifies the alignment between the RGB and alpha channels, providing a reliable measure of RGB-alpha consistency.
We conduct extensive experiments comparing our method against baseline techniques, including image inpainting combined with image matting~\cite{kim2024zim} and dichotomous image segmentation methods~\cite{zheng2024birefnet,Qin_U2UNet}. Experimental results demonstrate the superior performance of our approach in generating high-quality detail and RGB-alpha aligned content.

\section{Related Work}\label{sec:related_work}
\noindent{\bf Transparent Image and Video Generation.}
RGBA-related generation can be categorized into single-layer generation, multi-layered generation, and transparent video generation.
For single-layer generation, Text2Live~\cite{bar2022text2live} generates transparent edit layers by leveraging CLIP~\cite{radford2021clip} to construct an RGB editing loss.
LayerDiffuse~\cite{zhang2024transparent} introduces latent transparency for RGBA image generation, encoding the alpha channel into the pretrained Stable Diffusion’s latent space without affecting the decoder’s output, and decodes the alpha channel with a latent transparency decoder.
Zippo~\cite{xie2024zippo} compresses the RGB and alpha distributions into a single diffusion model, enabling RGB-to-alpha, alpha-to-RGB, and text-to-RGBA generation.
Alfie~\cite{quattrini2024alfie} and DiffMatting~\cite{hu2024diffumatting} generate images with a pure background first, then extract the alpha channel using GrabCut and image matting, respectively.
For multi-layered generation, Text2Layer~\cite{zhang2023text2layer} formulates a two-layer generation task and can generate transparent foreground and background at the same time.
LayerDiff~\cite{huang2024layerdiff} and ART~\cite{pu2025art} can generate multi-layered images in one diffusion denoising process.
TransPixar~\cite{wang2025transpixar} proposes the only transparent video generation method using sequence extension for the alpha channel in DiT~\cite{peebles2023DiT}'s self-attention.
While previous research has explored transparent image generation, our work extends inpainting capabilities to transparent images, introducing Trans-Adapter as a novel plug-and-play solution for diffusion-based inpainting.

\begin{figure*}[t]
  \centering
   \includegraphics[width=1.0\linewidth]{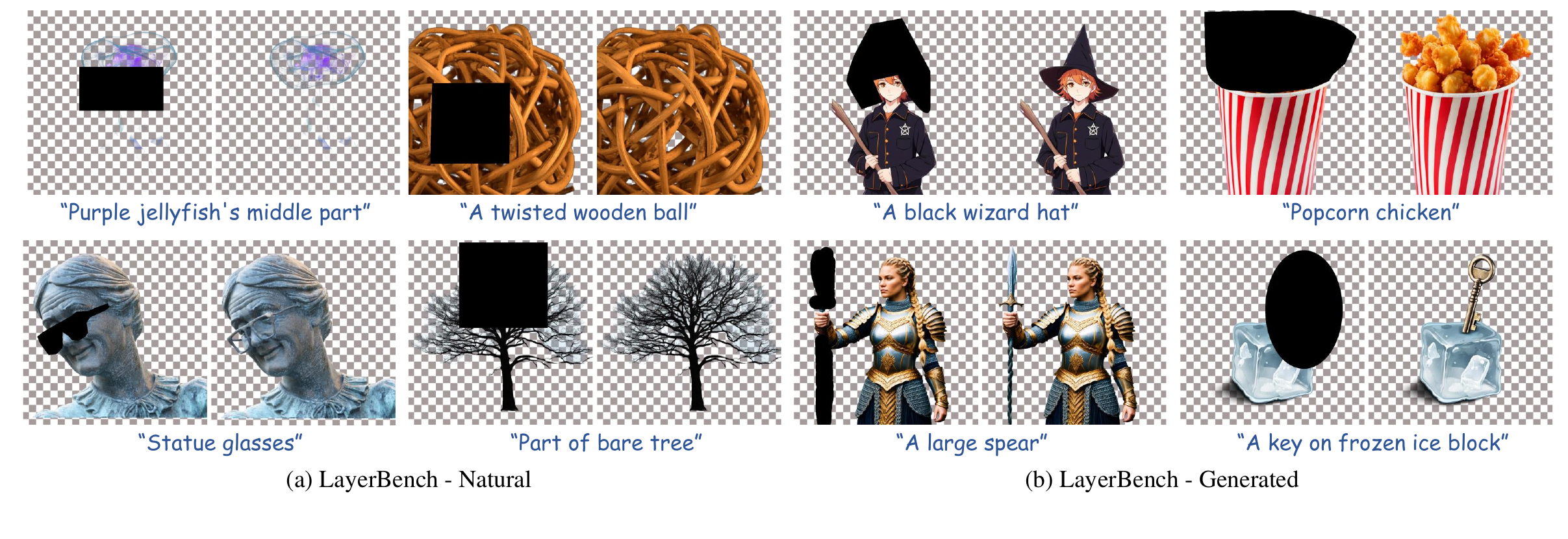}
   \vspace{-12mm}
   \caption{Sample images and their corresponding text prompts (mask-simple) from the proposed LayerBench benchmark. Our benchmark includes many challenging cases, such as images with complex alpha channel structures, transparent objects, and diverse artistic styles. Images shown in (b) are sourced from MAGICK~\cite{burgert2024magick} and DIM~\cite{xu2017deep} to complement the real portion (a) of our benchmark.}
   \vspace{-3.5mm}
   \label{fig:layerbench}
\end{figure*}

\noindent{\bf Image Inpainting.}
Image inpainting has been extensively studied in the past, focusing on the exploration of patch-matching techniques with different architectural choices ~\cite{telea2004image,barnes2009patchmatch,criminisi2004region} and GAN-based approaches~\cite{pathak2016gan1,yu2018gan2,reed2016gan3,xu2018attngan,zhang2017stackgan,zhang2018stackgan++}. Recent methods typically begin with pretrained diffusion models to leverage the exceptional generative capabilities.
RePaint~\cite{lugmayr2022repaint} introduces a training-free inpainting method based on DDPM~\cite{ho2020ddpm}, where unmasked regions are sampled and blended with the denoised masked region at each step of the reverse diffusion process.
Blended Latent Diffusion~\cite{avrahami2022blended,avrahami2023blended} extends this strategy to the latent space.
For finetuning-based approaches, Imagen Editor~\cite{wang2023imagen} is finetuned from Imagen~\cite{imagen}.
SmartBrush~\cite{xie2023smartbrush} is built on Stable Diffusion, incorporating mask boundary information to guide the denoising process for more intuitive editing.
Stable-Diffusion Inpainting~\cite{sd15} is the official finetuned version of Stable Diffusion, modifying the UNet~\cite{unet} by increasing its input channels to encode both the masked image and the alpha channel.
HD-Painter~\cite{manukyan2023hdpaint} and PowerPaint~\cite{powerpaint} further refine Stable-Diffusion Inpainting, improving generation quality and supporting multiple tasks independently.
ControlNet~\cite{zhang2023controlnet} enhances diffusion models by incorporating various conditioning signals, including inpainting.
Similarly, BrushNet~\cite{ju2024brushnet} and BrushEdit~\cite{li2024brushedit} introduce a plug-and-play branch that integrates with fine-tuned community models.
Building on BrushNet and ControlNet, MagicQuill~\cite{liu2024magicquill} enables intuitive content alterations by combining automatically estimated text prompts with user-drawn strokes.
While these methods offer powerful editing capabilities, they primarily operate on standard RGB images and lack support for RGBA editing, limiting their applicability in workflows requiring transparency handling.


\noindent{\bf Image Matting and Dichotomous Image Segmentation.}
Image matting and dichotomous image segmentation both aim to extract foreground objects with fine edge details.
Building on the success of SAM~\cite{kirillov2023segment} and its large-scale training, several recent works have explored its applicability to high-precision matting.
MattingAnything~\cite{li2024matting} follows SAM’s framework, supporting various prompt types—including box, point, and text—to enable more flexible matting.
ZIM~\cite{kim2024zim} enhances fine-grained image matting in a zero-shot setting, improving mask precision through a dataset from large-scale segmentation-to-matting conversion.
Dichotomous image segmentation methods BiRefNet~\cite{zheng2024birefnet} and $\text{U}^{2}$-Net~\cite{Qin_U2UNet} also focus on refining segmentation boundaries, contributing to high-quality foreground extraction. 
While matting and dichotomous segmentation focus on extracting precise foreground masks, our proposed method targets transparent image inpainting, ensuring transparency-aware restoration with aligned RGB and alpha content, which is a more challenging problem.

\section{LayerBench Benchmark}
\label{sec:dataset}
In this study, we introduce LayerBench, the first high-quality benchmark designed specifically for evaluating transparent image inpainting. To further enhance the evaluation framework, we propose a non-reference metric for evaluating alpha edge quality. We first review some existing transparent image datasets below before introducing our benchmark.

\noindent{\bf Existing Transparent Image Datasets.} Existing transparent image datasets remain limited in size and quality when compared to large-scale datasets like LAION~\cite{schuhmann2022laion} that contain billions of images. 
Among the available datasets, MAGICK~\cite{burgert2024magick} is the only one that balances both scale and quality. It contains 150K transparent images generated using SDXL~\cite{sdxl}, with chroma-keying applied to extract alpha maps.
In contrast, natural transparent images are significantly rarer than their generated counterparts. 
For instance, DIM~\cite{xu2017deep} contains only 431 foreground transparent images, while the Semantic Matting Dataset~\cite{sun2021semantic} extends DIM to 726 images. Distinctions-646~\cite{qiao2020attention} consists of 646 foreground images.
In the video matting domain, VideoMatte240K~\cite{lin2021real} provides 240K unique transparent frames extracted from 484 high-resolution videos, with a primary focus on human figures.
Recently, MULAN~\cite{tudosiu2024mulan} introduces a multi-layer dataset featuring 44K layers derived from matting results of LAION~\cite{schuhmann2022laion} and COCO~\cite{lin2014coco}, but the resolution and quality of the foreground images remain relatively low.
%
%
None of the existing data is specifically designed for high-quality transparent image inpainting evaluation.

\subsection{LayerBench}
Some examples in LayerBench are shown in Fig.~\ref{fig:layerbench}. To cover different use cases, LayerBench is composed of two subsets: LayerBench-Natural and LayerBench-Generated, each containing 400 images. The LayerBench-Natural subset consists of natural images sourced from online PNG stocks and matting datasets~\cite{xu2017deep}.
For the LayerBench-Generated subset, we select 200 images with high Aesthetic Score~\cite{schuhmann2022laion} from the MAGICK dataset~\cite{burgert2024magick} that are exclusive for testing. 
The remaining 200 images are generated using LayerDiffusion applied to SDXL~\cite{sdxl} and the community model RealVisXL-V4.0, with text prompts generated by ChatGPT.
To simulate real-world use cases, we manually annotate masks for each image, marking the regions that require inpainting. Our dataset includes all common mask types, such as strokes, regular shapes, regions masked by Bézier curves, and object masks.
Besides, similar to EditBench~\cite{wang2023imagen}, we provide full-text descriptions for the entire image and mask-simple annotations to describe the high-level content within the masked regions. The full-text descriptions are generated using LLaVA~\cite{liu2023llava}, following the same procedure as our training set's text prompt labeling. 
For mask-simple annotations, we manually label the semantic information of the masked regions.
All images in LayerBench have a resolution of 1024$\times$1024.

\begin{figure}[t]
    \centering
    \includegraphics[width=\linewidth]{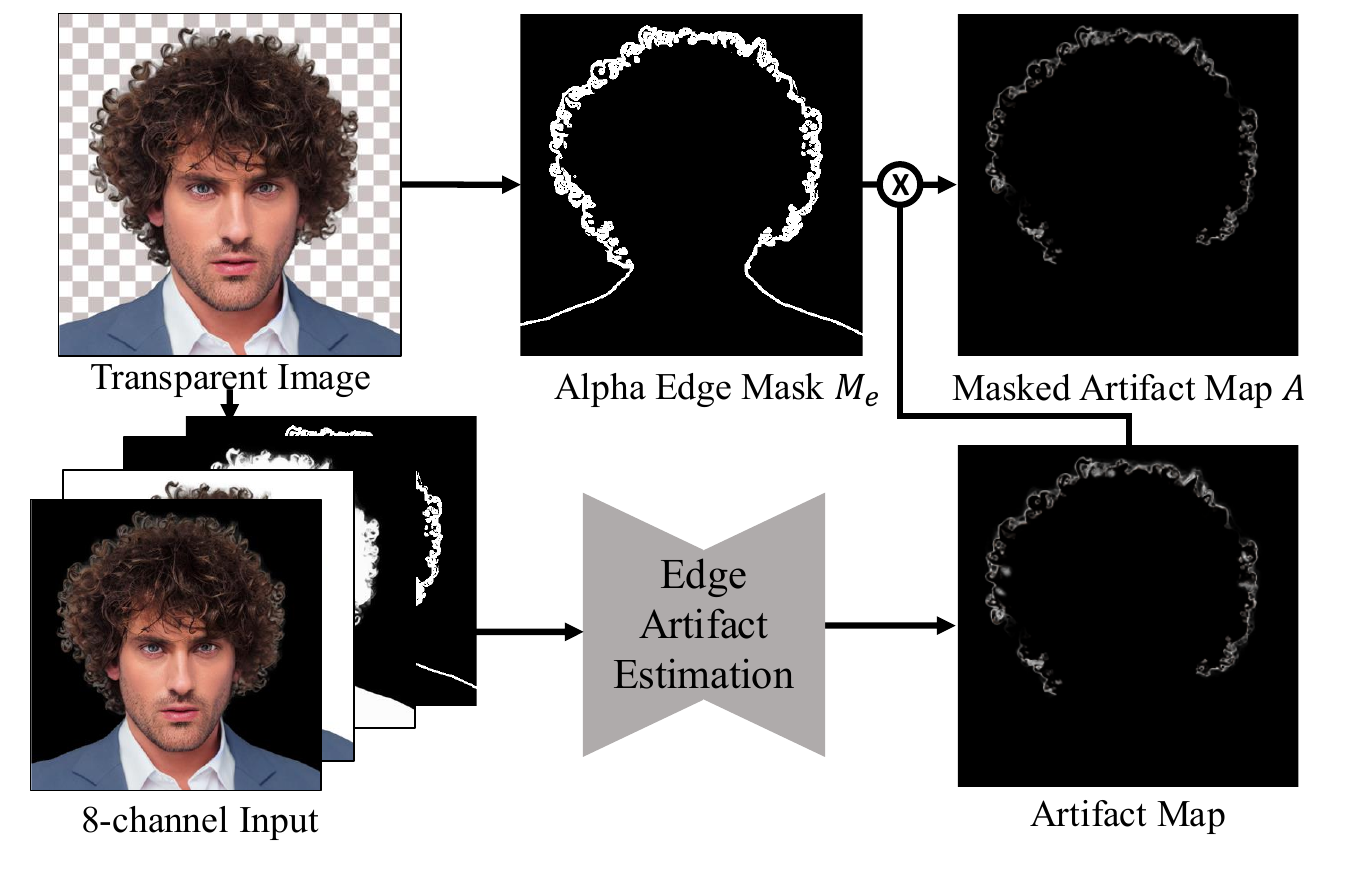}
    \vspace{-10mm}
    \caption{Overview of the Alpha Edge Quality (AEQ) assessment pipeline. The classifier receives an eight-channel input, consisting of the transparent image composited on white and black backgrounds, the alpha channel, and the alpha edge mask. It outputs a probability map that represents low-quality edge regions.}
    \label{fig:aeq_pipeline}
    \vspace{-5mm}
\end{figure}

\subsection{Alpha Edge Quality Assessment}
\label{subsec:aeq}
As discussed in Sec.~\ref{sec:intro}, misalignment between the RGB and alpha channels in transparent images can cause jagged edges, degrading visual quality during compositing.
To measure the alignment quality of RGB and alpha channels, we propose a new non-reference metric named Alpha Edge Quality (AEQ), which can be conveniently computed using a specially trained classifier.

The classifier is a lightweight Convolutional Neural Network (CNN) that performs binary segmentation, \ie, distinguishing low-quality edge regions from the high-quality ones.
It takes an eight-channel input formed by the transparent image layered on white and black backgrounds, then concatenates with the alpha channel $\alpha$ and alpha edge mask $M_e$. We use backgrounds of two different colors to better reveal the quality of alpha edges.
The AEQ can be computed as one minus the average low-quality edge probabilities within the evaluation mask $\mathcal{M}_e$, which surrounds the edges of the alpha map.
Specifically, the AEQ is defined as:
\begin{equation}
\mathrm{AEQ} = 1 - \frac{1}{|\mathcal{M}_e|} \sum_{(x,y) \in \mathcal{M}_e} \mathcal{F}(I_w, I_b, \alpha,\mathcal{M}_e)_{x,y}
\end{equation}
where $\mathcal{F}$ is the classifier, the evaluation mask $\mathcal{M}_e$ is obtained by applying the Canny edge detector to the alpha map with a threshold of 20, followed by a dilation operation with a kernel size of five pixels.  
Note that we only consider $\mathcal{M}_e$ within the inpainting mask.
AEQ ranges between 0 and 1, where higher values indicate better quality.
The classifier $\mathcal{F}$ is trained to be robust in discerning various alpha edge qualities. This is achieved by training it carefully with an extensive set of augmented data as potential input, starting from a set of high-quality transparent images. More details about the training of the AEQ classifier are provided in the supplementary material.

\begin{figure*}[t]
  \centering
   \includegraphics[width=1.0\linewidth]{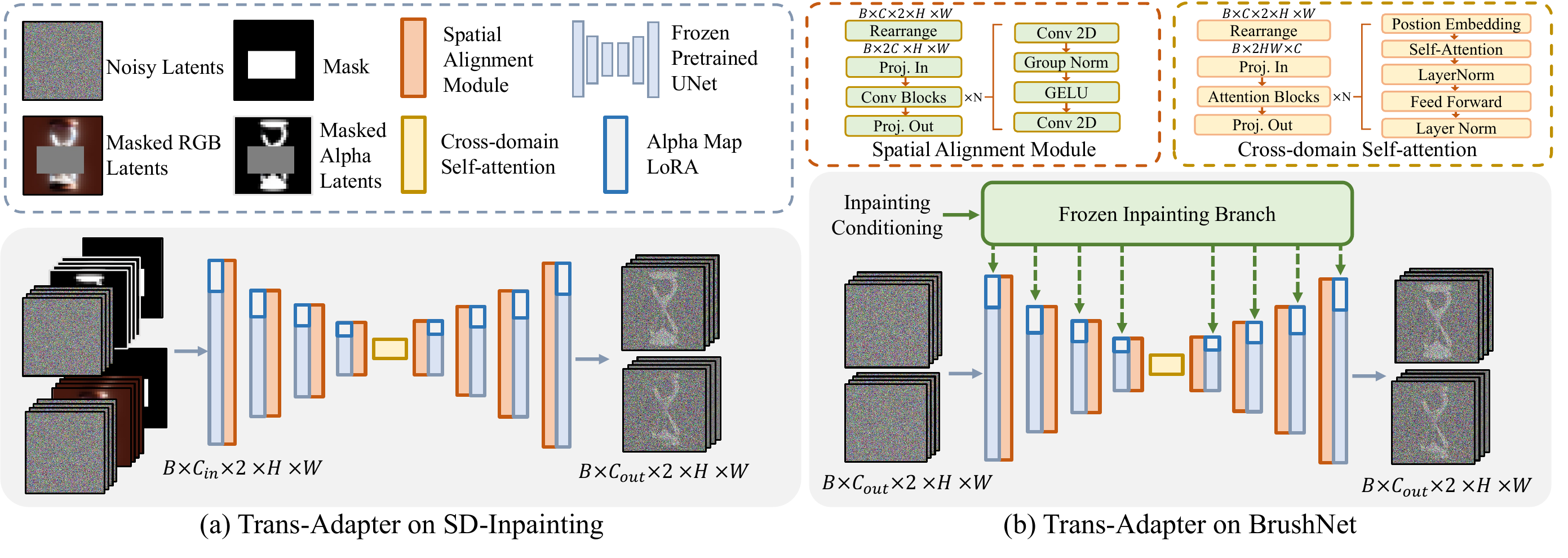}
   \vspace{-8mm}
   \caption{Trans-Adapter can be flexibly integrated into different image inpainting frameworks. We demonstrate two instantiations: (a) SD-Inpainting~\cite{sd15}, which expands the input channels to encode the additional mask and masked image, and (b) BrushNet~\cite{ju2024brushnet}, which introduces an inpainting branch that accepts the inpainting conditioning. In BrushNet, the inpainting conditioning is the same as the SD-Inpainting's input. Trans-Adapter enables these pipelines to process transparent images by incorporating trainable cross-domain self-attention, spatial alignment modules, and alpha map LoRA.}
   \vspace{-4.5mm}
   \label{fig:trans-adapter}
\end{figure*}

\section{Trans-Adapter}
\label{sec:method}
Trans-Adapter is designed as a plug-and-play module that can be seamlessly integrated into existing diffusion-based inpainting frameworks. 
In this study, we show two instantiations of this adaptation, as illustrated in Fig.~\ref{fig:trans-adapter}.
The first instantiation expands the input channels of Stable Diffusion's U-Net to encode the mask and masked image, followed by fine-tuning the diffusion model. This approach is used in methods such as SD-Inpainting, with official implementations available for SD1.5~\cite{sd15}, SDXL~\cite{sdxl}, and Flux~\cite{flux}.
The second instantiation introduces a trainable inpainting branch, as seen in ControlNet~\cite{zhang2023controlnet} and BrushNet~\cite{ju2024brushnet}.  

Having a good network design can be tricky as the network needs to simultaneously generate both the RGB image and its corresponding alpha map, while enforcing feature alignment between color and transparency information.
We overcome this challenge by using an inflated network design. Specifically, we decompose the RGBA image into a separate padded RGB image~\cite{zhang2024transparent} and its alpha channel, treating them as a two-frame video with a 5D input tensor, where the feature map is denoted as $f \in \mathbb{R}^{b\times c\times 2 \times h \times w}$.  
As the tensor passes through the original weights of a pre-trained diffusion model, the latent is reshaped into $ f_r \in \mathbb{R}^{2b\times c \times h \times w}$, stacking the latents of the RGB and alpha channels along the batch size dimension $ b $.
This design allows the model to capture cross-channel dependencies and avoid independent processing biases that can lead to RGB-alpha misalignment.  

To enforce spatial coherence between the RGB and alpha components, we add a spatial alignment module and cross-domain self-attention in our design, inspired by AnimateDiff~\cite{guoanimatediff}. In the two instantiations shown in Fig.~\ref{fig:trans-adapter}, the original network parameters remain frozen, requiring only the newly introduced spatial alignment module and cross-domain self-attention to be trained. To facilitate the fine-tuning of these modules, all output projection layers in these modules are initialized to zero, a design choice validated by ControlNet~\cite{zhang2023controlnet}, ensuring a stable training process while preserving the pre-trained model's capabilities.  

\vspace{0.1cm}
\noindent{\bf Training Strategy.}
We employ a two-stage training strategy. In the first stage, the Alpha Map LoRA~\cite{hu2022lora} is trained to guide the model in reconstructing the alpha map.
In the second stage, we jointly fine-tune the spatial alignment module, cross-domain self-attention module, and the alpha map LoRA, which primarily enables the model to inpaint well-aligned RGBA content. More training details are provided in the supplementary material.

\vspace{0.1cm}
\noindent{\bf Alpha Map LoRA.}
We introduce a LoRA~\cite{hu2022lora} module into the pre-trained diffusion model to equip it with the ability to inpaint the alpha map. In this stage, the original weights of UNet remain frozen, and only the LoRA module attached to the U-Net is trained. The LoRA weights are initialized to zero, ensuring that the module learns only the necessary residuals for alpha map generation. During training, we simultaneously select both the RGB content and the alpha map from transparent images for fine-tuning. 
To further enhance the model's ability to distinguish between RGB content and alpha content, we use different text prompts for each during training. Specifically, for alpha map inpainting, we prepend the text prompt with ``alpha map of'', while for RGB content, we use the original prompt. This explicit separation guides the model to learn distinct inpainting strategies for RGB images and alpha maps.

\vspace{0.1cm}
\noindent{\bf Spatial Alignment Module.}
In the shallower layers of Stable Diffusion's U-Net, we incorporate a spatial alignment module. Specifically, we utilize convolution layers to enforce spatial synchronization between the alpha map and the RGB latent representation at corresponding spatial locations. This design ensures that the RGB and alpha maps remain aligned even after passing through the VAE decoder. Given a feature tensor $f \in \mathbb{R}^{b\times c\times 2 \times h \times w}$, we first reshape it into $f_r \in \mathbb{R}^{b\times 2c \times h \times w}$ and then use:
\begin{equation}
f = f_r+ \mathcal{Z}_c(\mathbf{ConvBlock}(f_r)),
\end{equation}
where $\mathcal{Z}_c$ denotes the zero-initialized convolution layer.

\vspace{0.1cm}
\noindent{\bf Cross-domain Self-Attention.}
To ensure the consistency of the masked alpha channel with its surrounding regions, we incorporate cross-domain self-attention at U-Net's bottleneck. For high-frequency details in the alpha channel that are difficult to predict, such as hair, cross-domain self-attention allows the masked region to better reference its surrounding areas for inpainting. After adding a 2D positional embedding to the latent representation, it is then reshaped into $f_r \in \mathbb{R}^{b\times 2hw \times c}$ and processed through:
\begin{equation}
\textbf{self-attention}(f_r) = \text{softmax}\left(\frac{\mathbf{Q}_i\mathbf{K}_i}{\sqrt{D}} \right)\mathbf{V}_i,
\end{equation}
where $\mathbf{Q}_i$, $\mathbf{K}_i$, and $\mathbf{V}_i$ represent ${f}_r$ processed by MLPs for query, key, and value. Finally, a zero-initialized MLP $\mathcal{Z}_M$ will be applied to the feature:
\begin{equation}
f = f_r+ \mathcal{Z}_M(\textbf{AttentionBlock}(f_r)) .
\end{equation}

\begin{figure*}[t]
  \centering
   \includegraphics[width=1.0\linewidth]{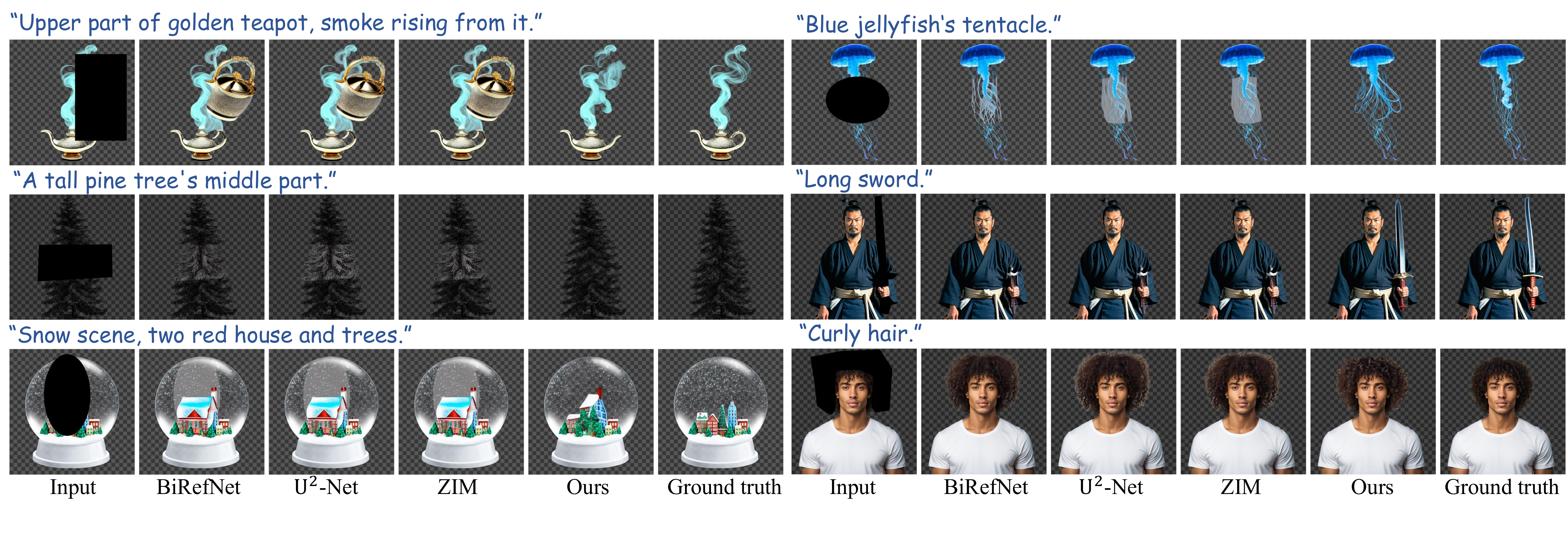}
   \vspace{-11mm}
   \caption{Visual comparison of Trans-Adapter with official SD1.5-inpainting combined with other image matting and dichotomous image segmentation methods. Since our approach simultaneously generates both the alpha map and RGB, it performs well even for complex details and transparent objects.}
   \label{fig:comparison}
   \vspace{-2mm}
\end{figure*}

\noindent\textbf{Training Loss.}
We adopt the vanilla training objective as proposed in DDPM~\cite{ho2020ddpm}, which can be expressed as:
\begin{equation}
    \mathcal{L} = \mathbb{E}_{\mathcal{E}(x_0), y, \epsilon\sim\mathcal{N}(0, \mathit{I}), t}\left\lbrack 
\lVert \epsilon - \epsilon_\theta(z_t, t, \tau_\theta(y),C) \rVert_2^2 \right\rbrack ,
\end{equation}
where $y$ means the textual prompt, $\tau_\theta(\cdot)$ is a text encoder mapping the prompt to a sequence of tokens. The conditioning information is denoted as $C$, which, in the context of image inpainting, consists of the masked image latent and the mask. In the training stage of a latent diffusion network, an input image $x_0$ is initially encoded into the latent space by a frozen encoder, yielding $z_0 = \mathcal{E}(x_0)$.

\begin{table*}[]
\caption{
Quantitative comparison of Trans-Adapter with different foreground extraction methods. For BiRefNet~\cite{zheng2024birefnet}, we use the checkpoint from RMBG 2.0 for better results. The metrics are calculated on the entire inpainted image, which adopts a mask blending with the original image. All experiments with SD1.5 are conducted at a resolution of $512\times512$, while those with SDXL are done at a resolution of $1024\times1024$.
} \label{tab:comparison}
\vspace{-3mm}
\resizebox{1.0\textwidth}{!}{
\begin{tabular}{lllccccccccl}
\hline
Inpainting Strategy                           &                &  & \multicolumn{4}{c}{Pure Noise}   & \multicolumn{1}{l}{} & \multicolumn{4}{c}{Blended Noise}                     \\ \hline
Base Model                                         &    Method            &  & AS$\uparrow$   & LPIPS$\downarrow$ & CLIP Sim$\uparrow$ & AEQ$\uparrow$ &                      & AS$\uparrow$   & LPIPS$\downarrow$ & CLIP Sim$\uparrow$ & \multicolumn{1}{c}{AEQ$\uparrow$} \\ \cline{1-2} \cline{4-7} \cline{9-12} 
                           & Input ($512\times512$)          &  & 6.157 & /      & 27.143    &  0.9866  &                      & 6.157 & /      & 27.143    & 0.9866   \\  \cline{1-2} \cline{4-7} \cline{9-12} 
\multirow{4}{*}{SD1.5-inpaint}                     & ZIM~\cite{kim2024zim}     &  & 6.011 & 0.0697 & \textbf{27.061}    & 0.9866   &                      & 6.044 & 0.0526 & 27.040    &      0.9874                  \\
                           & $\text{U}^2$-Net~\cite{Qin_U2UNet}        &  & 5.970 & 0.0720 & 26.989    &  0.9502  &                      & 6.007 & 0.0560 & 26.978    &    0.9537                    \\
                           & BiRefNet~\cite{zheng2024birefnet}       &  & \textbf{6.027} & 0.0682 & 27.034    &  \textbf{0.9889}   &                     & 6.055 & 0.0515 & \textbf{27.049}    &        \textbf{0.9886}                \\
                           & Ours           &  & 6.025 & \textbf{0.0591} & 26.870    &  0.9871   &                      & \textbf{6.097} & \textbf{0.0408} & 27.030    &     0.9878                    \\ \cline{1-2} \cline{4-7} \cline{9-12} 
\multirow{4}{*}{SD1.5-BrushNet}                    & ZIM~\cite{kim2024zim}     &  & 5.929 & 0.0764 & 26.906    &  0.9853  &                      & 5.942 & 0.0601 & \textbf{26.970}    &          0.9852              \\
                           & $\text{U}^2$-Net~\cite{Qin_U2UNet}        &  & 5.898 & 0.0779 & 26.813    & 0.9549   &                      & 5.897 & 0.0640 & 26.892    &   0.9547                     \\
                           & BiRefNet~\cite{zheng2024birefnet}       &  & 5.946 & \textbf{0.0742} & 26.897    &  0.9844  &                      & 5.954 & 0.0592 & 26.952    &    0.9863                    \\
                           & Ours           &  & \textbf{6.021} & 0.0757 & \textbf{26.987}    &   \textbf{0.9856}  &                     & \textbf{6.053} & \textbf{0.0505} & 26.941    &    \textbf{0.9869}                     \\ \cline{1-2} \cline{4-7} \cline{9-12} 
                           & Input ($1024\times1024$)          &  & 6.181 & /      & 27.153    & 0.9849   &                      & 6.181 & /      & 27.153    & 0.9849   \\
                          
                           \cline{1-2} \cline{4-7} \cline{9-12} 
SDXL                                               & LayerDiffusion &  & 5.851 & 0.0903 & 26.831    & 0.9760   &                      & 6.016 & 0.0642 & 27.097    &       0.9781                 \\
                           \cline{1-2} \cline{4-7} \cline{9-12}                                                    
\multirow{4}{*}{SDXL-inpaint}                      & ZIM~\cite{kim2024zim}     &  & 6.075 & 0.0693 & 27.126    & 0.9794   &                      & 6.115 & 0.0461 & 27.111    &        0.9828                \\
                           & $\text{U}^2$-Net~\cite{Qin_U2UNet}        &  & 6.031 & 0.0708 & 27.002    & 0.8093   &                      & 6.066 & 0.0500 & 27.038    &   0.8367                     \\
                           & BiRefNet~\cite{zheng2024birefnet}       &  & \textbf{6.087} & 0.0681 & 27.076    &  0.9843  &                      & 6.129 & 0.0453 & 27.104    &                0.9859        \\
                           & Ours           &  & 6.042 & \textbf{0.0632} & \textbf{27.332}    &    \textbf{0.9844} &                      & \textbf{6.140} & \textbf{0.0434} & \textbf{27.134}    &     \textbf{0.9872}                    \\
                           \hline
\end{tabular}}
\vspace{-5mm}
\end{table*}

\noindent\textbf{Training Data.}
Existing transparent image datasets primarily originate from matting datasets~\cite{xu2017deep,sun2021semantic,yu2018gan2} and generated images~\cite{burgert2024magick}.
Although matting datasets are natural images of high quality, their limited quantity makes them unsuitable for large-scale training.
To ensure our method performs well on both natural and generated transparent images, we collect a new dataset by purchasing transparent images from an online PNG stock and manually filtering out those with jagged edges, blurriness, or low resolution (long side $<$ 600px), resulting in 35K high-quality images.
Each image is then paired with a text prompt generated by the image captioning model LLaVA~\cite{liu2023llava}. 
This training dataset encompasses a diverse range of categories, including people, objects, buildings, hand-drawn cartoon scenes, cartoon characters, and commercial artwork elements. 
Finally, we merge it with a subset of MAGICK~\cite{burgert2024magick}, selecting 90\% for training and reserving 10\% for benchmarking.

\section{Experiments}
\label{sec:experiment}

\begin{figure*}[t]
  \centering
  \resizebox{0.95\textwidth}{!}{
   \includegraphics[width=0.9\linewidth]{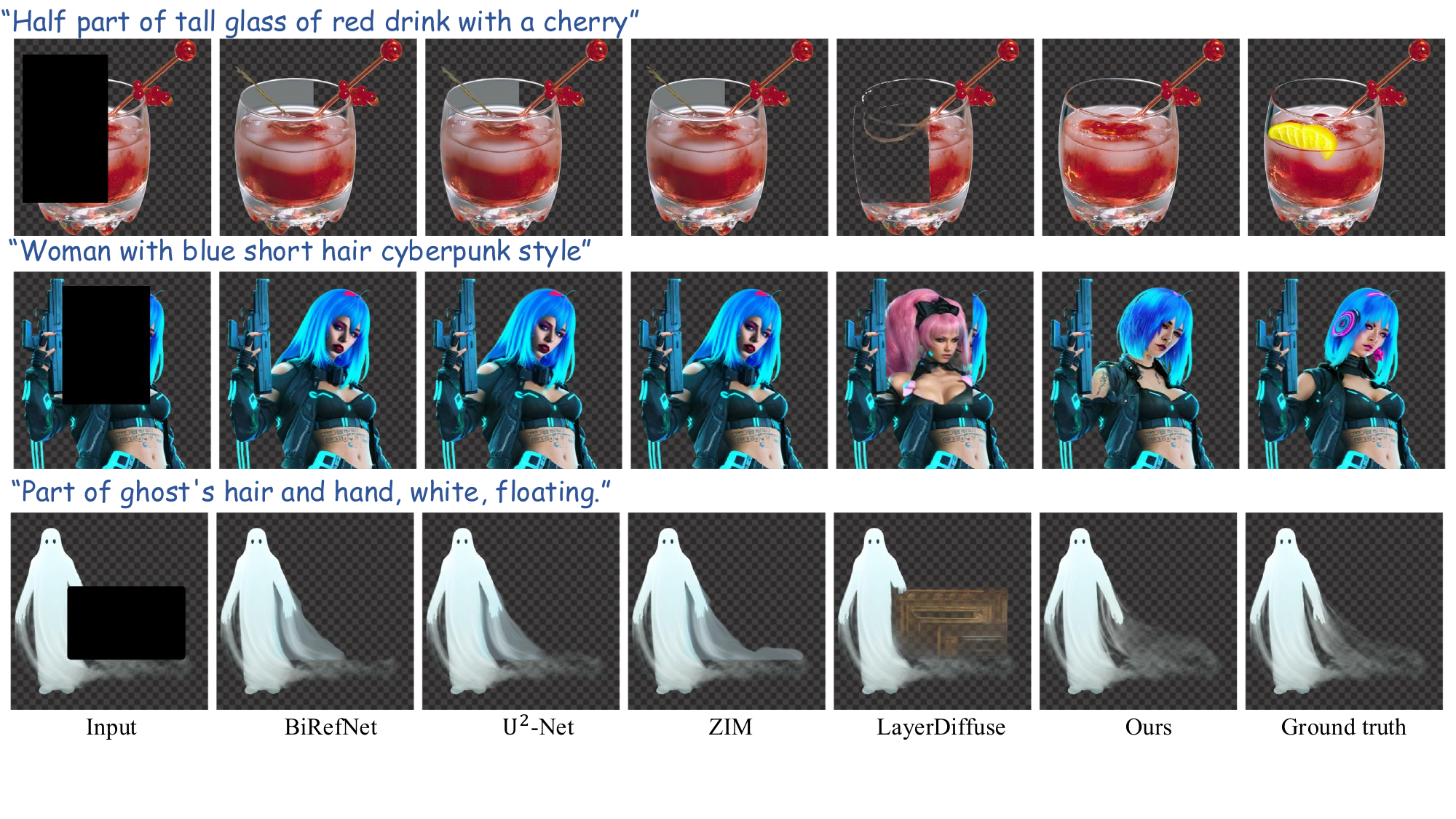}   
 }
   \vspace{-10mm}
   \caption{Visual comparison of Trans-Adapter's performance on SDXL-Inpainting against the official SDXL-Inpainting method, combined with the image matting method, ZIM~\cite{kim2024zim}, and dichotomous image segmentation methods $\text{U}^2$-Net~\cite{qin2020u2} and BiRefNet~\cite{zheng2024birefnet}. The proposed method generates alpha maps aligned with the RGB images while preserving their quality.}
   \label{fig:comparison_sdxl}
   \vspace{-4mm}
\end{figure*}

To demonstrate the generalizability of our approach, we trained Trans-Adapter on SD1.5, BrushNet based on SD1.5, and SDXL. To ensure fairness in subsequent experiments, we maintained the same training resolution as in pretraining: 512$\times$512 for SD1.5 and BrushNet, and 1024$\times$1024 for SDXL. SD1.5 was trained on an Nvidia 3090, while SDXL was trained on an H100 due to memory constraints. 
Detailed settings are provided in the supplementary material.

\subsection{Comparison with Existing Methods}
Apart from the proposed AEQ (Sec.~\ref{subsec:aeq}), following Ju \etal \cite{ju2024brushnet}, we also use Aesthetic Score (AS)~\cite{schuhmann2022laion}, LPIPS~\cite{zhang2018lpips}, and CLIP Similarity~\cite{radford2021clip} to assess image quality, masked region preservation, and text alignment, respectively.
As these metrics are designed for RGB images, they cannot be directly applied to RGBA images.
Besides, since transparent images may contain arbitrary values in the RGB channels when the alpha value is zero, direct comparison using the RGB channels of transparent images is not reliable.
To better evaluate the quality of our transparent image inpainting, we 
composite the transparent image onto both white and black backgrounds to generate two RGB images. We then compute the metrics against the original image processed in the same manner and take the average of the two results as the final score.
This approach effectively addresses the issue where certain transparent images with defined RGB values (\eg, line drawings created in black) cannot be properly evaluated when blended with backgrounds of the same RGB value.
During the evaluation, we replace the unmasked regions with the original input to ensure our metric only focuses on the inpainting region.

Since there is no existing study about transparent image inpainting, we mainly compare our method with different diffusion-based inpainting methods followed by image matting~\cite{kim2024zim} or dichotomous image segmentation~\cite{zheng2024birefnet,Qin_U2UNet}.
First, we composite a transparent image onto a gray background. After inpainting the target regions using~\cite{sd15,sdxl,ju2024brushnet}, we apply image matting and dichotomous image segmentation techniques to extract the edited content.
To keep a fair comparison, we choose the same inpainting network to compare our method with the previous two-stage pipelines, as shown in Table~\ref{tab:comparison}.
To provide a more comprehensive evaluation, we assess each method under two commonly used inpainting strategies: (1) initializing with pure noise, and (2) initializing with blended noise (where the masked region is blended with noise at a strength of 0.99). These strategies reflect typical settings in practical inpainting and editing scenarios.

The results demonstrate that our method achieves competitive and often better results than previous two-stage pipelines.
As shown in Fig.~\ref{fig:comparison}, compared with these two-stage pipelines, our method can better reconstruct details in the alpha channel and avoid misalignment between RGB and alpha channels.
Since LayerDiffuse~\cite{zhang2024transparent} can also inpaint the transparent image by utilizing Blended Latent Diffusion~\cite{avrahami2022blended}'s denoising strategy, we compare our method with it in Fig.~\ref{fig:comparison_sdxl}. As can be observed, our approach achieves more accurate structure reconstruction and better boundary consistency in challenging regions.

\begin{table*}[]
\caption{
Quantitative comparison of training with different datasets, network structures, and methods. `localized' means localized inpainting -- we only inpaint the pixels around the edge of the alpha channel. All these experiments are conducted on the SD1.5-inpainting model with a resolution of 512.
} \label{tab:ablation}
\vspace{-2mm}
\resizebox{1.0\textwidth}{!}{
\begin{tabular}{lllccccccccc}
\hline
                             &                                           &  & \multicolumn{4}{c}{Pure Noise}       & \multicolumn{1}{l}{} & \multicolumn{4}{c}{Blended Noise}    \\ \hline
Type                         & Method                                    &  & AS↑   & LPIPS↓ & CLIP Sim.↑ & AEQ↑   &                      & AS↑   & LPIPS↓ & CLIP Sim.↑ & AEQ↑   \\ \cline{1-2} \cline{4-7} \cline{9-12} 
                         & Ours                                          &  & 6.025 & 0.0591 & 26.870     & 0.9871 &                      & 6.097 & 0.0408 & 27.030     & 0.9878 \\ \cline{1-2} \cline{4-7} \cline{9-12} 
\multirow{2}{*}{Dataset}     & w/o MAGICK                                &  & 5.992 & 0.0633 & 26.934     & 0.9867 &                      & 6.073 & 0.0435 & 27.037     & 0.9873 \\
                             & w/o Ours                                  &  & 6.007 & 0.0613 & 26.921     & 0.9865 &                      & 6.067 & 0.0457 & 27.071     & 0.9881 \\ \cline{1-2} \cline{4-7} \cline{9-12} 
\multirow{3}{*}{Network}     & AnimateDiff                               &  & 5.989 & 0.0623 & 26.793     & 0.9859 &                      & 6.067 & 0.0459 & 27.032     & 0.9872 \\
                             & w/o Spatial Align.                        &  & 5.542 & 0.0712 & 26.170     & 0.9881 &                      & 5.685 & 0.0580 & 26.642     & 0.9897 \\
                             & w/o Self-Attn                             &  & 6.011 & 0.0621 & 26.720     & 0.9872 &                      & 6.065 & 0.0461 & 26.987     & 0.9880 \\ \cline{1-2} \cline{4-7} \cline{9-12} 
\multirow{3}{*}{RGB Padding} & Telea~\etal~\cite{telea2004image}           &  & 6.017 & 0.0603 & 26.470     & 0.9871 &                      & 6.070 & 0.0427 & 26.681     & 0.9853 \\
                             & Telea~\etal~\cite{telea2004image} localized &  & 6.021 & 0.0627 & 26.865     & 0.9868 &                      & 6.064 & 0.0457 & 27.005     & 0.9861 \\
                             & Grey Background                           &  & 6.031 & 0.0584 & 27.017     & 0.9647 &                      & 6.103 & 0.0462 & 26.998     & 0.9717 \\ \cline{1-7} \cline{9-12} 
\end{tabular}}
\vspace{-3mm}
\end{table*}

\subsection{Ablation Study}
\noindent{\bf Training Set.} 
We analyze the effect of training data composition by removing either the MAGICK or our dataset. As shown in Table~\ref{tab:ablation}, removing either dataset leads to a slight decrease in AS, LPIPS, and AEQ, demonstrating that both subsets contribute to the overall performance. This highlights the significance of our newly introduced dataset, as each subset brings complementary benefits to the transparent image inpainting model.

\begin{figure}[t]
  \centering
   \includegraphics[width=1.0\linewidth]{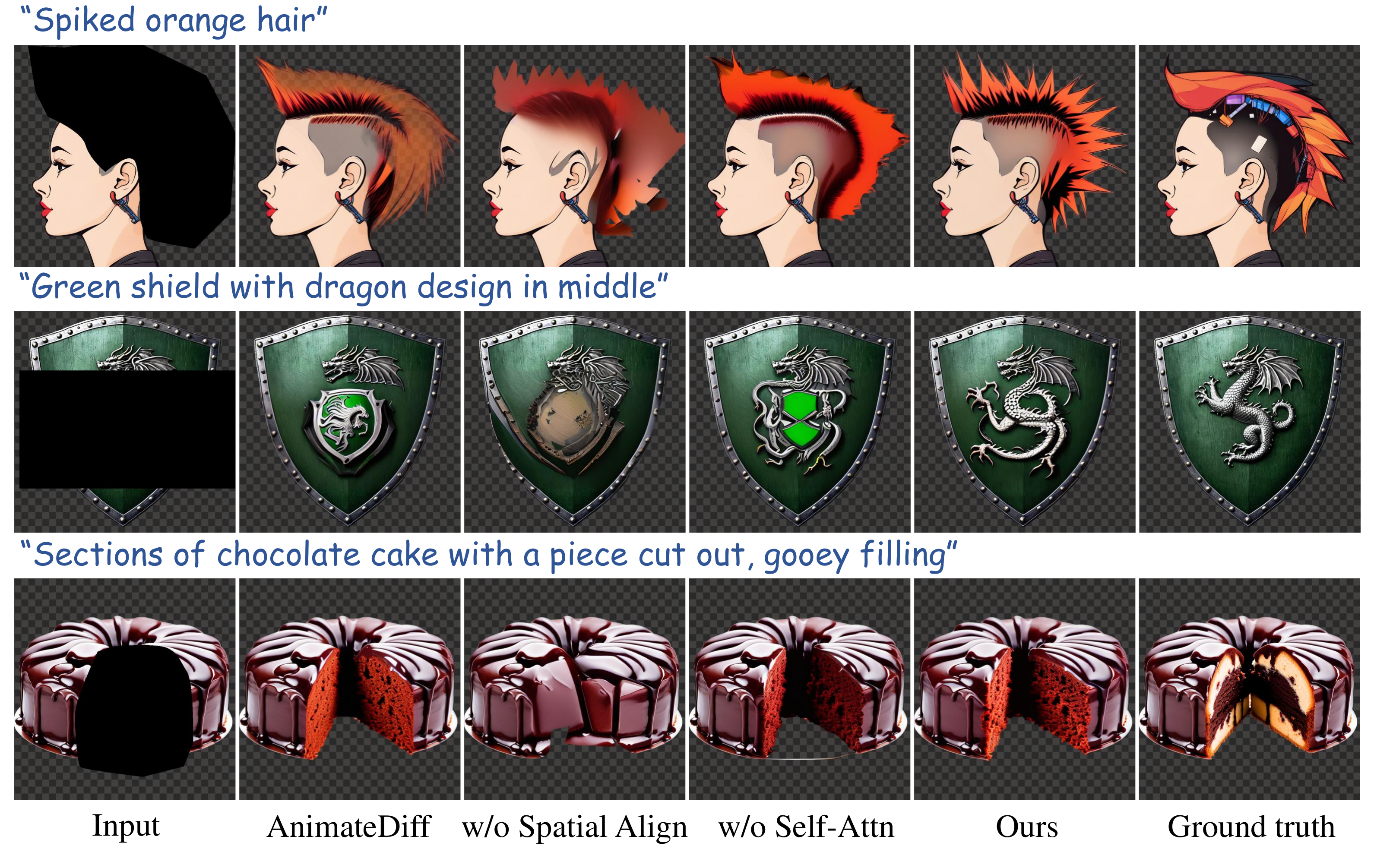}
   \vspace{-8mm}
   \caption{Training Trans-Adapter with different network structures. The details of the hair in the first row demonstrate the effectiveness of our spatial alignment module. 
 As shown in the second and third rows, the absence of self-attention leads to a lack of global information, making artifacts more likely to appear in the image for Animate Diff and our method w/o Self-Attn, which ultimately affects the aesthetic metrics.
 }
   \vspace{-3.5mm}
   \label{fig:ablation_network}
\end{figure}

\noindent{\bf Network Structure.} 
We also conduct experiments where we replace all spatial alignment modules with cross-domain self-attention (w/o Spatial Align.) and replace all the cross-domain self-attention modules with the spatial alignment modules (w/o Self-Attn). Additionally, we train the original AnimateDiff for comparison, as presented in Table~\ref{tab:comparison}. The visual results shown in Fig.~\ref{fig:ablation_network} illustrate the importance of our designed modules. Due to the removal of spatial alignment, the image occasionally exhibits large pure-color edges originating from the background, which unexpectedly increases AEQ.

\begin{figure}[t]
  \centering
   \includegraphics[width=0.9\linewidth]{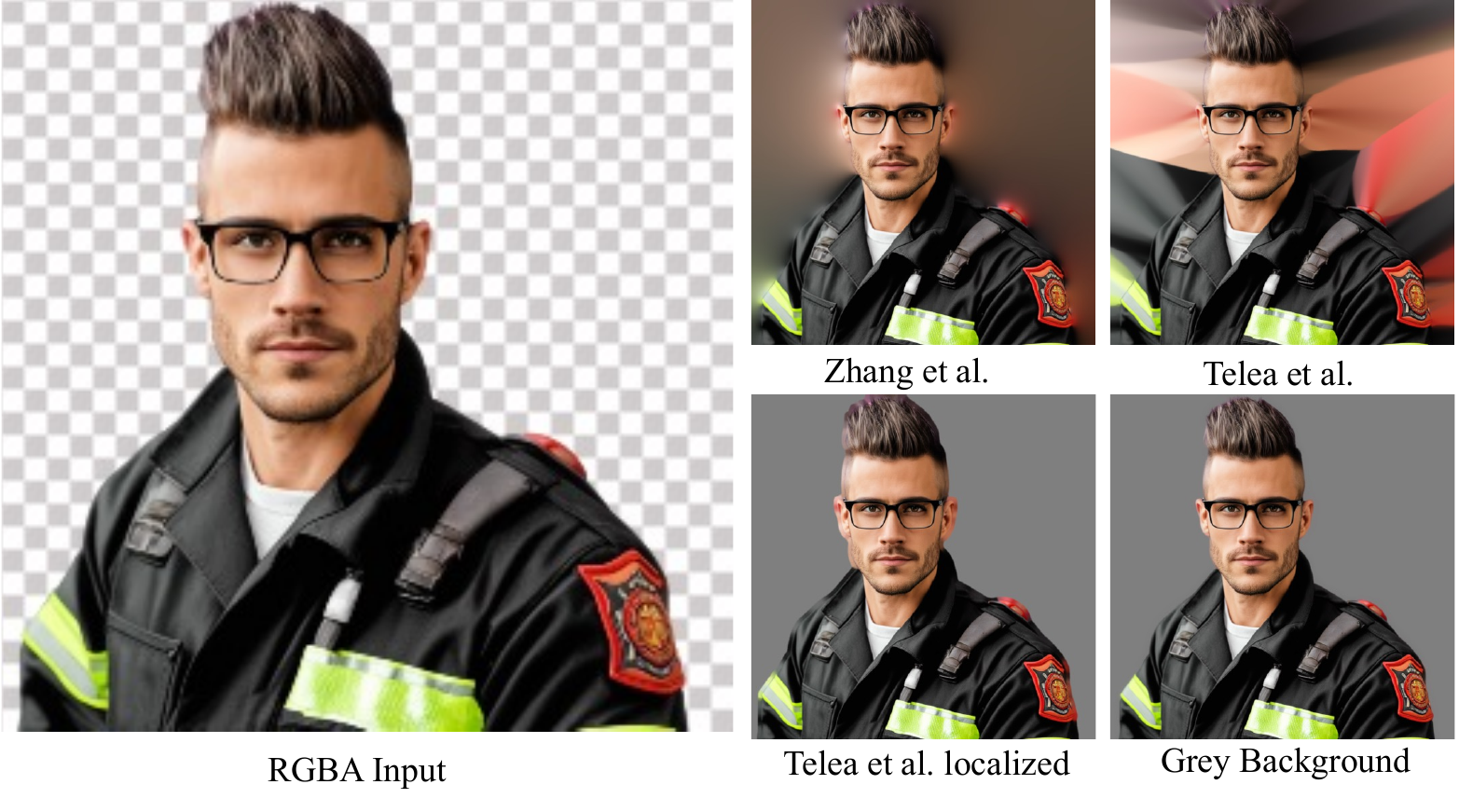}
   \vspace{-4mm}
  \caption{Visual comparison of various RGB padding strategies for filling transparent regions.}  \vspace{-7mm}
   \label{fig:rgb_padding}
\end{figure}

\noindent{\bf RGB Padding Method.}
Due to the nature of diffusion models, achieving perfectly aligned RGB and alpha channels is challenging.
Since the RGB values in regions where alpha is 0 can be arbitrarily chosen, selecting an appropriate RGB padding strategy ensures that even if the alignment between RGB and alpha is imperfect, the final edge quality remains unaffected.
However, as the network also needs to predict the RGB values in 0-alpha regions, an overly artificial padding strategy may cause the adapter to focus excessively on reconstructing the transparent regions, ultimately degrading the final image quality.
An effective RGB padding strategy should ensure that, after training, the algorithm preserves both edge quality and the aesthetic appeal of the final image.
As shown in Fig.~\ref{fig:rgb_padding}, we evaluate different RGB padding methods for regions where the alpha value is lower than 20, including Zhang~\etal~\cite{zhang2024transparent}'s RGB padding, Telea~\etal~\cite{telea2004image}'s inpainting, and grey background padding.
Additionally, we test expanding the alpha map by 30 pixels, inpainting only these expanded regions using Telea~\etal~\cite{telea2004image}’s inpainting method (the inpainting method applied in OpenCV), and setting the remaining areas to gray. This approach is denoted as Telea~\etal~\cite{telea2004image} localized in Table~\ref{tab:ablation}.
We select LayerDiffuse~\cite{zhang2024transparent}'s RGB padding as our baseline method, as it strikes a balance between aesthetic quality and edge preservation, as shown in Table~\ref{tab:ablation}.

\section{Conclusion}
\label{sec:conclusion}
We have introduced Trans-Adapter, a plug-and-play framework that enables diffusion-based inpainting models to process transparent images directly. Trans-Adapter also supports controllable editing via ControlNet and integrates seamlessly into various community models.
To facilitate the rigorous evaluation of transparent image inpainting methods, we proposed LayerBench, a benchmark specifically designed for this task. Furthermore, we introduced a novel non-reference evaluation metric that quantifies the alignment between the RGB and alpha channels, providing a reliable measure of RGB-alpha consistency.
Experimental results demonstrate that Trans-Adapter outperforms existing pipelines in preserving transparency consistency and producing high-quality inpainted results. 

\vspace{-4mm}
\paragraph{Acknowledgement:}
This study is supported under the RIE2020 Industry Alignment Fund Industry Collaboration Projects (IAF-ICP) Funding Initiative, as well as cash and in-kind contribution from the industry partner(s).
{
    \small
    \bibliographystyle{ieeenat_fullname}
    \bibliography{main}
}

\clearpage
\setcounter{page}{1}
\maketitlesupplementary
\setcounter{section}{0} 
\renewcommand{\thesection}{\Alph{section}}

\section{Potential Applications}
\begin{figure}[h]
  \centering
   \includegraphics[width=1.0\linewidth]{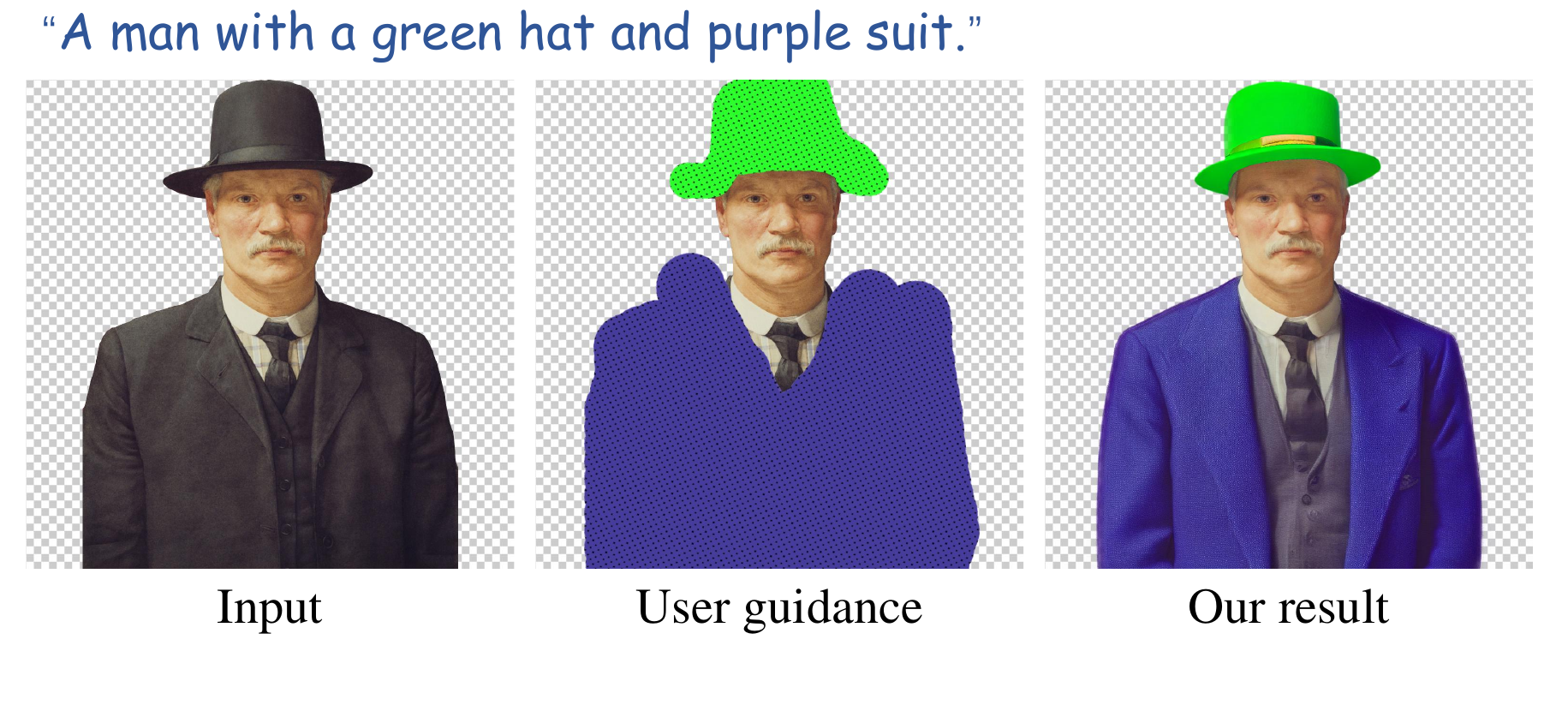}
   \vspace{-10mm}
   \caption{Demonstration of color-stroke-based transparent image editing. Users can draw on a transparent image and obtain an estimated result based on their strokes and a provided text prompt.}
   \vspace{-2mm}
   \label{fig:color_stroke}
\end{figure}

\noindent\textbf{Transparent Image Editing}.
By not inpainting from the pure noise, our method can also serve as a transparent image editor naturally. 
As shown in Fig.~\ref{fig:color_stroke}, users can plot color strokes on the original transparent image. 
These color strokes will be considered as a mask.
Then, we add noise to the drawn RGB and alpha map with a strength of 0.99 as the initial noise.
Finally, we perform denoising following the previous approach to obtain the edited result.

\begin{figure}[h]
  \centering
   \includegraphics[width=1.0\linewidth]{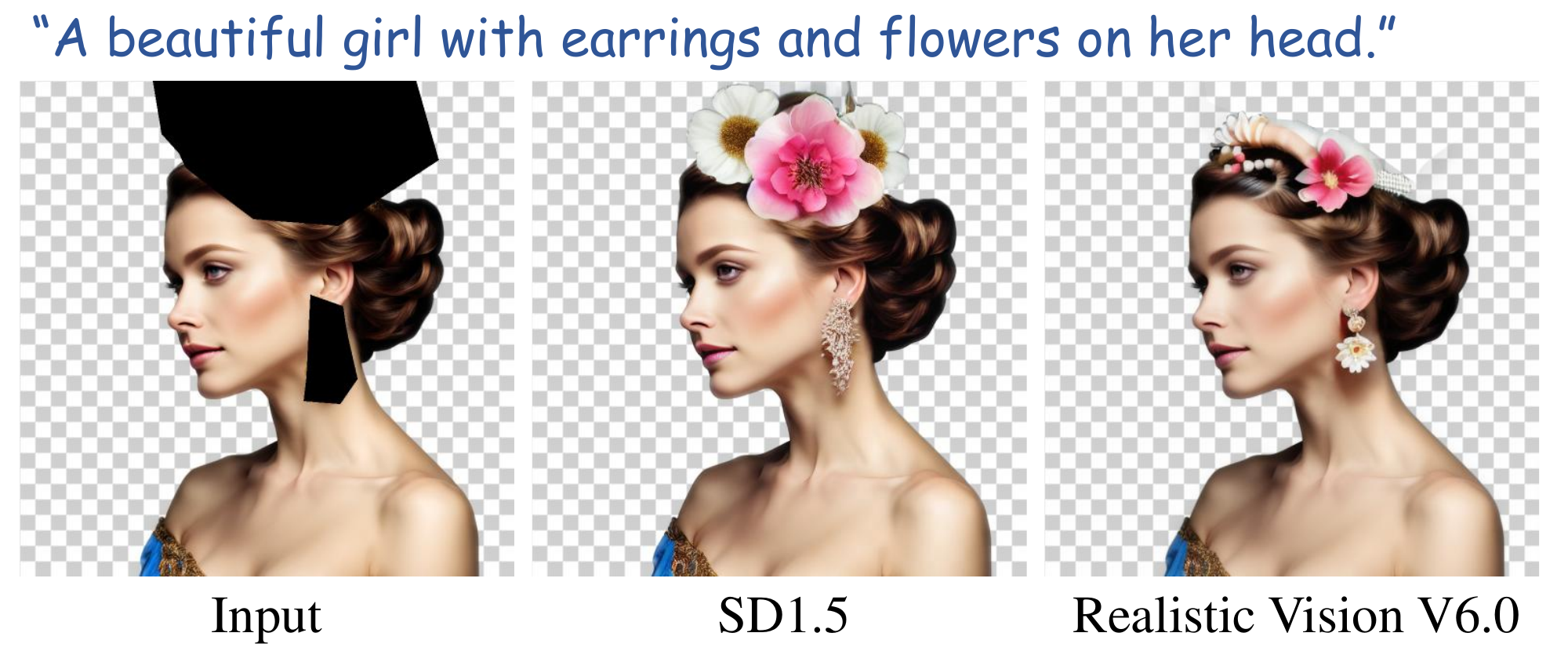}
   \vspace{-7mm}
   \caption{While combining with BrushNet, our Trans-Adapter also supports community models for inpainting. }
   \vspace{-4mm}
   \label{fig:comminity model}
\end{figure}

\noindent\textbf{Extending to Community Models.}
Since both BrushNet and our Trans-Adapter are plug-and-play modules, they can be applied together to other community models, enabling inpainting based on different models. As shown in the Fig.~\ref{fig:comminity model}, by using different community models, we can achieve inpainting effects in various styles.

\begin{table*}[t]
\begin{center}
\caption{Quantitative comparison of different LoRA training strategies under pure noise and blended noise settings. }\label{tab:ablation_lora}
\resizebox{1.0\textwidth}{!}{
\begin{tabular}{llccccccccc}
\hline
                           &                      & \multicolumn{4}{c}{Pure Noise}      & \multicolumn{1}{l}{} & \multicolumn{4}{c}{Blended Noise}   \\ \hline
Method                     &                      & AS↑   & LPIPS↓ & CLIP Sim↑ & AEQ↑   &                      & AS↑   & LPIPS↓ & CLIP Sim↑ & AEQ↑   \\ \cline{1-1} \cline{3-6} \cline{8-11} 
Ours                       & \multicolumn{1}{r}{} & \textbf{6.025} & \textbf{0.0591} & 26.870    & \textbf{0.9871} &                      & \textbf{6.097} & 0.0408 & 27.030    & \textbf{0.9878} \\ \cline{1-1} \cline{3-6} \cline{8-11} 
Frame-specific LoRA        &                      & 5.992 & 0.0616 & 27.165    & 0.9855 &                      & 6.089 & 0.0416 & 27.036    & 0.9863 \\
Frozen LoRA                &                      & 5.979 & 0.0637 & 27.043    & 0.9817 &                      & 6.067 & 0.0471 & 27.005    & 0.9831 \\
Frozen Frame-specific LoRA &                      & 5.985 & 0.0624 & 27.166    & 0.9856 &                      & 6.082 & \textbf{0.0408} & 27.023    & 0.9859 \\
w/o LoRA                   &                      & 5.982 & 0.0769 & \textbf{27.363}    & 0.9863 &                      & 6.091 & 0.0430 & \textbf{27.103}    & 0.9868 \\ \hline
\end{tabular}
}
\vspace{-8mm}
\end{center}
\end{table*}

\begin{figure}[h]
  \centering
   \includegraphics[width=1.0\linewidth]{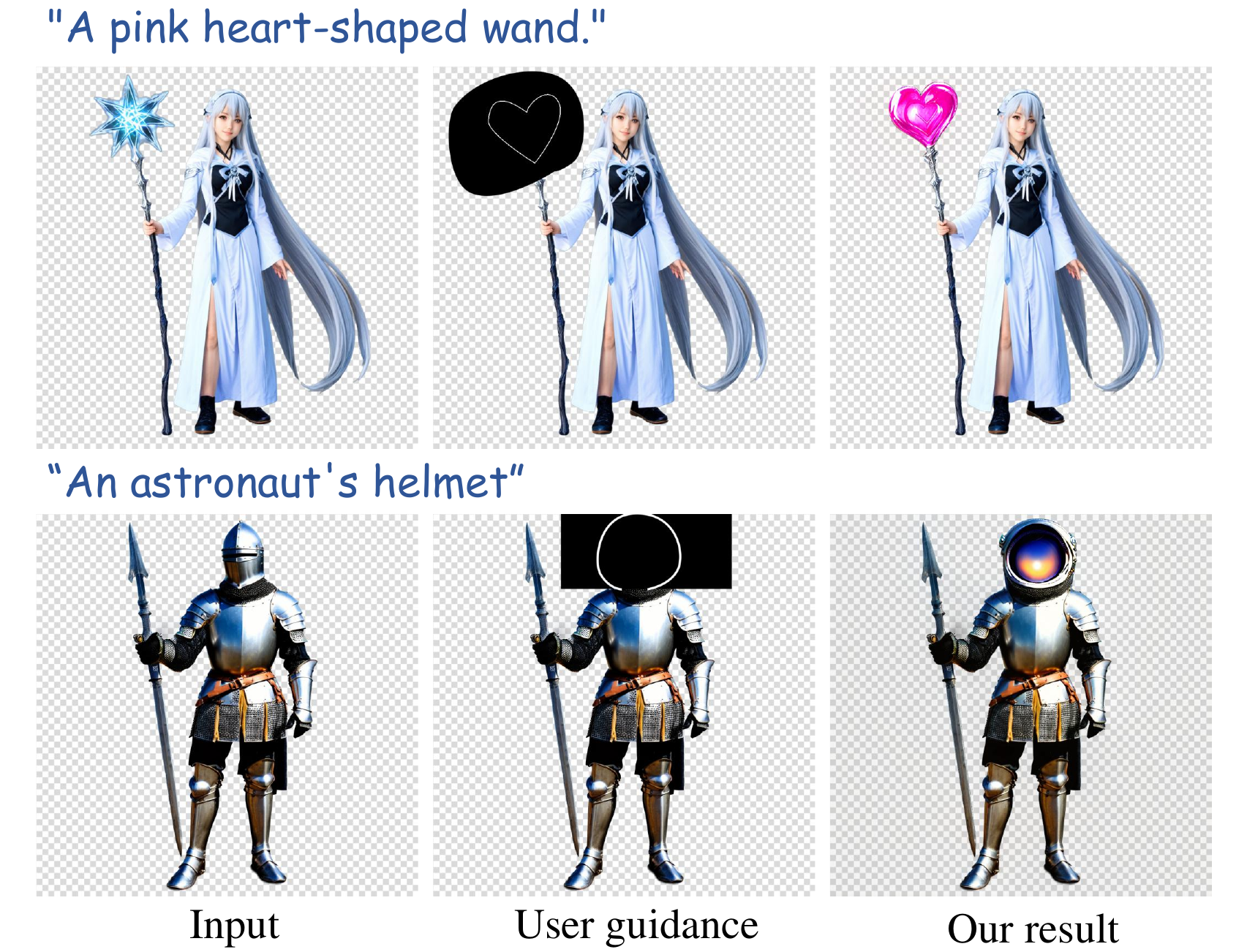}
   \vspace{-6mm}
   \caption{Our approach can be directly combined with control models like ControlNet~\cite{zhang2023controlnet} to enhance functionality. Users can define a mask and outline the inpainting region to generate a transparent image.}
   \vspace{-3.5mm}
   \label{fig:controlnet}
\end{figure}

\noindent\textbf{ControlNet Extension.}
As shown in Fig.~\ref{fig:controlnet}, we demonstrate that existing control models like ControlNet~\cite{zhang2023controlnet} can be applied to our model for enriched functionality. 
Since ControlNet does not provide a control model for SD-Inpainting and only supports T2I generation, we apply Trans-Adapter to BrushNet based on SD1.5 and then use ControlNet (with Scribbles) to control image details.
The visualization results demonstrate that our method can effectively integrate ControlNet to control inpainting outputs, allowing users to guide structure and details more precisely.

\section{More Details of AEQ Assessment}

\noindent{\bf Data Augmentation.}
We collect a high-quality transparent image dataset with clean alpha edges with 1,000 images from the online PNG stock and matting data.
Starting from these transparent images, we simulate images with varying degrees of edge quality to create a comprehensive training set. To generate low-quality edges, we:
(1) set regions with an alpha value lower than 50 to a solid color, then perform a dilation operation or a Gaussian blur operation on the alpha map. For the expanded region at the alpha map, we multiply it with the mask generated by the fractal noise to create a more realistic edge degradation effect.
(2) composite the images with different backgrounds (solid color backgrounds + natural scenery backgrounds) and then process these images using different segmentation and matting
methods~\cite{kirillov2023segment,li2024matting}. 
To quantify edge degradation, we compute the difference between the original image's alpha map and its augmented counterpart, identifying regions with significant discrepancies as low-quality edges.
In this way, we can train a segmentation network for alpha edge quality assessment.

\noindent{\bf Training Loss.}
Since the edge quality assessment is a binary classification task, we use a weighted cross-entropy loss to address class imbalance, as low-quality edge pixels are much fewer than high-quality ones. We assign a higher weight to the low-quality class and further emphasize edge regions using an edge mask $\mathcal{M}_e$.

Here, $I_{\text{concat}}$ denotes the input, $\mathcal{F}$ is the segmentation network, $y$ is the ground truth label map, and $\mathcal{M}_e$ is the edge mask. The loss is defined as:
\begin{equation}
\mathcal{L} = \frac{1}{HW} \sum_{i =1}^{HW} \mathrm{CE}(\mathcal{F}(I_{\text{concat}})_i, y_i, w_{y_i}) \cdot (1 + w_e \mathcal{M}_{e})
\end{equation}
where $w_{y_i}$ denotes the class weight ($w_0$ for high-quality, $w_1$ for low-quality, typically $w_0$ as 1.0, $w_1$ as 10.0), and $w_e$ is the edge weight (set to 4.0). This loss encourages the model to focus more on low-quality edge regions during training.

\begin{figure}[t]
  \centering
    \includegraphics[width=1.0\linewidth]{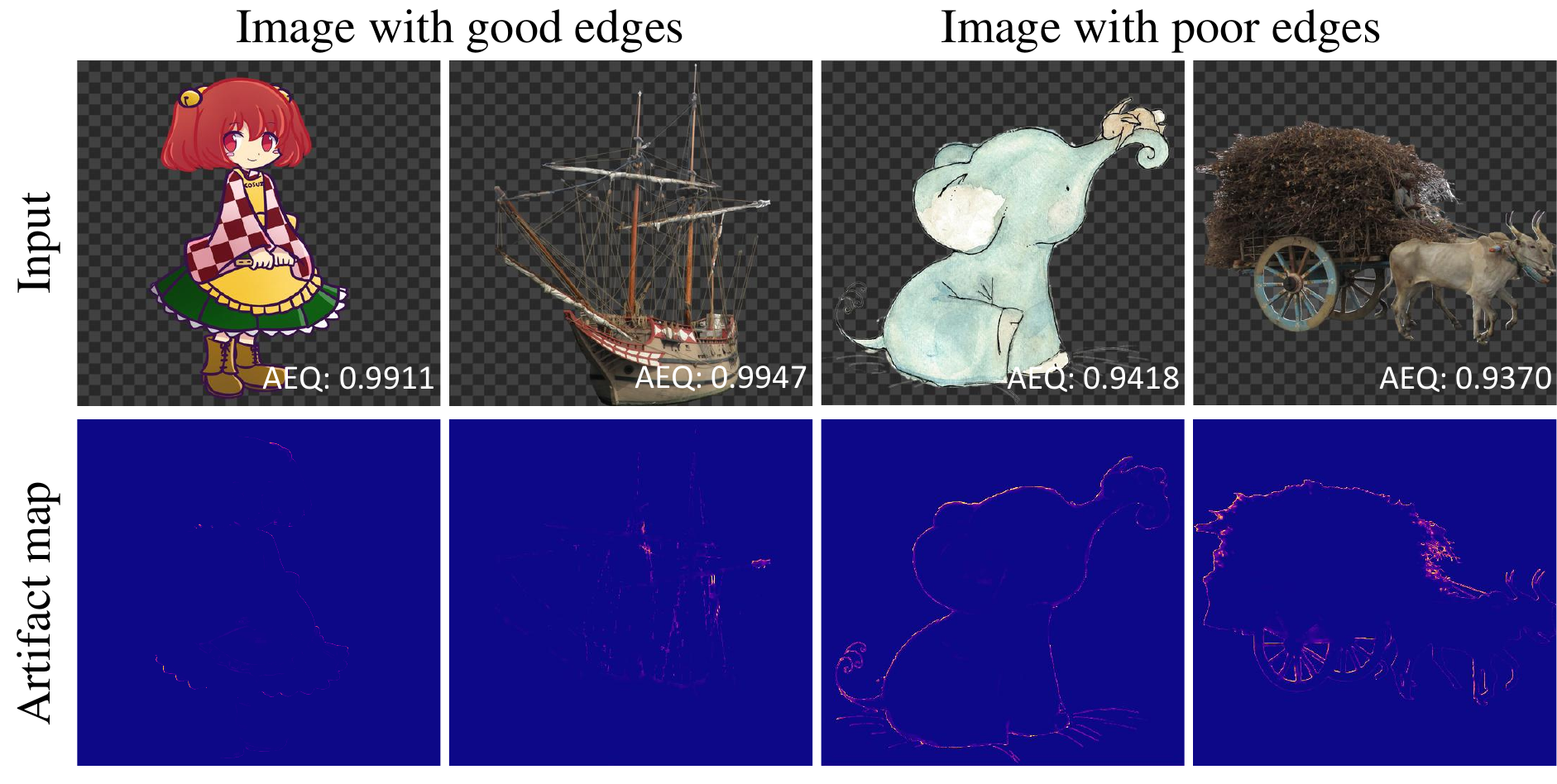}
   \vspace{-6mm}
   \caption{Estimated artifact map of our proposed AEQ for images in different styles. (Zoom in for better visualization.)}
   \vspace{-5mm}
   \label{fig:aeq_vis}
\end{figure}

\noindent{\bf AEQ Visualization.}
To demonstrate the effect of our proposed AEQ, we visualize the estimated artifact maps for images in various styles. As shown in Fig.~\ref{fig:aeq_vis}, AEQ effectively highlights boundary artifacts across different image types, indicating its strong generalization ability beyond specific styles.

\noindent{\bf Network Architecture and Training Details.}
We adopt a lightweight U-Net-based segmentation network comprising three downsampling and three upsampling blocks. Each block contains two convolutional layers with ReLU activation and batch normalization. The network takes an 8-channel input and produces a 2-channel output, with hidden channel dimensions of 64, 128, and 256, respectively. We use the Adam optimizer with a learning rate of $1 \times 10^{-4}$. Training is performed on images with resolutions of $512 \times 512$ and $1024 \times 1024$, using a batch size of 4 and randomly selecting a resolution for each batch. The network is trained for 40{,}000 iterations.

\section{\bf More Details of Trans-Adapter}
\noindent{\bf Stage 1: Alpha Map LoRA Training.}
To enable the generation of large areas of pure black and pure white in the alpha map during inpainting, we adopt offset noise~\cite{offset_noise} with a weight of 0.1 during LoRA training, as conventional finetuning often struggles with such cases. The LoRA rank is set to 16 and the LoRA alpha is set to 32. We use the AdamW optimizer with an initial learning rate of $1 \times 10^{-4}$, and train for 40,000 steps with a batch size of 4.

\noindent{\bf Stage 2: Joint Finetuning.}
In the joint finetuning stage, we first load the pretrained LoRA weights, then zero-initialize the spatial alignment module and cross-domain self-attention module. These two modules are finetuned with a learning rate of $5 \times 10^{-5}$ using the AdamW optimizer for 100,000 steps, with a batch size of 4.
We conduct an ablation study to compare different training strategies, including frame-specific LoRA (training LoRA only on alpha images, affecting only the alpha channel), frozen LoRA (training LoRA on both RGB and alpha images, then freezing LoRA during stage 2), frozen frame-specific LoRA, and training without LoRA (directly training stage 2 without LoRA). As shown in Table~\ref{tab:ablation_lora}, our full method achieves the best overall performance in terms of AS, LPIPS, and AEQ metrics under both pure noise and blended noise settings. 
Our method achieves the best AEQ and LPIPS, validating the effectiveness of the two-stage training. Removing or freezing LoRA degrades AEQ and LPIPS, highlighting the importance of LoRA-based alpha map pretraining.

\section{Limitations}
\noindent
While our proposed method achieves strong performance in transparent image inpainting and editing, several limitations remain.
First, our approach is based on SD1.5 and SDXL, which are known to struggle with generating realistic faces and hands during inpainting. This can result in artifacts when editing or restoring these regions, limiting applicability in scenarios requiring high-fidelity human features.
Second, when the strength is set to 1.0 for SDXL-inpainting 0.1, the quality of the generated image degrades, which also affects our RGBA inpainting results. This limitation originates from the base checkpoint.
Finally, the alpha maps in the MAGICK dataset~\cite{burgert2024magick} used for training are not perfect; some regions, such as eyes, are translucent. As a result, our model may also produce undesired translucent areas in these regions during inpainting.

\end{document}